\documentclass{article} 
\usepackage[submission]{colm2025_conference}
\usepackage{subcaption}
\usepackage{amsmath}
\usepackage{amssymb}
\usepackage{array}
\usepackage{booktabs}
\usepackage{CJKutf8}
\usepackage{microtype}
\usepackage{hyperref}
\usepackage{url}
\usepackage{booktabs}
\usepackage{tcolorbox}
\usepackage{lineno}
\usepackage{graphicx}
\usepackage{hyperref}
\usepackage{microtype}
\usepackage{multirow} 
\usepackage{hyperref}
\tcbuselibrary{skins}
\definecolor{darkblue}{rgb}{0, 0, 0.5}
\hypersetup{colorlinks=true, citecolor=darkblue, linkcolor=darkblue, urlcolor=darkblue}
\tcbset{  
  aiboxc/.style={
    width=\columnwidth,
    colback=blue!6!white,
    colframe=black,
    colbacktitle=black,
    enhanced,
    center,
    attach boxed title to top left={yshift=-0.1in,xshift=0.15in},
    boxed title style={boxrule=0pt,colframe=white,},
  }
}
\newtcolorbox{AIboxC}[2][]{aiboxc,title=#2,#1}

\colmpreprinttrue

\title{VisTW: Benchmarking Vision-Language Models for\\ Traditional Chinese in Taiwan}
\author{Zhi Rui Tam\\
Independent Researcher\\
\AND
Ya-Ting Pai\thanks{Equal contribution}\\
University of Illinois Urbana-Champaign \\
\AND
Yen-Wei Lee\footnotemark[1]\\
Independent Researcher\\
\AND
Yun-Nung Chen\\
National Taiwan University
}

\begin{document}


\maketitle

\begin{abstract}
In this paper, we propose a comprehensive evaluation benchmark for Visual Language Models (VLM) in Traditional Chinese. Our evaluation suite, the first of its kind, contains two complementary components: (1) VisTW-MCQ, a collection of manually curated exam multi-choice questions from 21 academic subjects designed to test the broad knowledge and reasoning capabilities of VLMs; and (2) VisTW-Dialogue, an open dialogue benchmark comprising 131 image-question pairs manually created to evaluate VLMs' ability in free-form dialogue generation within Taiwanese cultural contexts. These benchmarks address a critical gap in the evaluation landscape, where existing benchmarks predominantly focus on English or Simplified Chinese, neglecting the unique linguistic and cultural aspects of Traditional Chinese used in regions like Taiwan and Hong Kong. Our analysis reveals significant performance differences across various VLMs and highlights specific challenges in processing Traditional Chinese visual content.
\footnote{Our code and dataset can be found in \url{https://github.com/TMMMU-Benchmark/evaluation} } 
\end{abstract}

\section{Introduction}

Vision-Language Models (VLMs) have achieved remarkable success in tasks such as image captioning, visual question answering (VQA), and cross-modal retrieval \citep{chen2022pali, li2022blip, faysse2024colpali}. However, current multimodal benchmarks predominantly focus on English, and when extended to Chinese, they primarily employ Simplified Chinese scripts common in mainland China \citep{das2024exams, winata2024worldcuisines, wang2024cvlue}. Traditional Chinese, widely used in Taiwan and Hong Kong, remains significantly underrepresented. Consequently, VLM capabilities in Traditional Chinese contexts are largely untested and unknown.

\begin{figure}
    \centering
    \includegraphics[width=1.0\columnwidth]{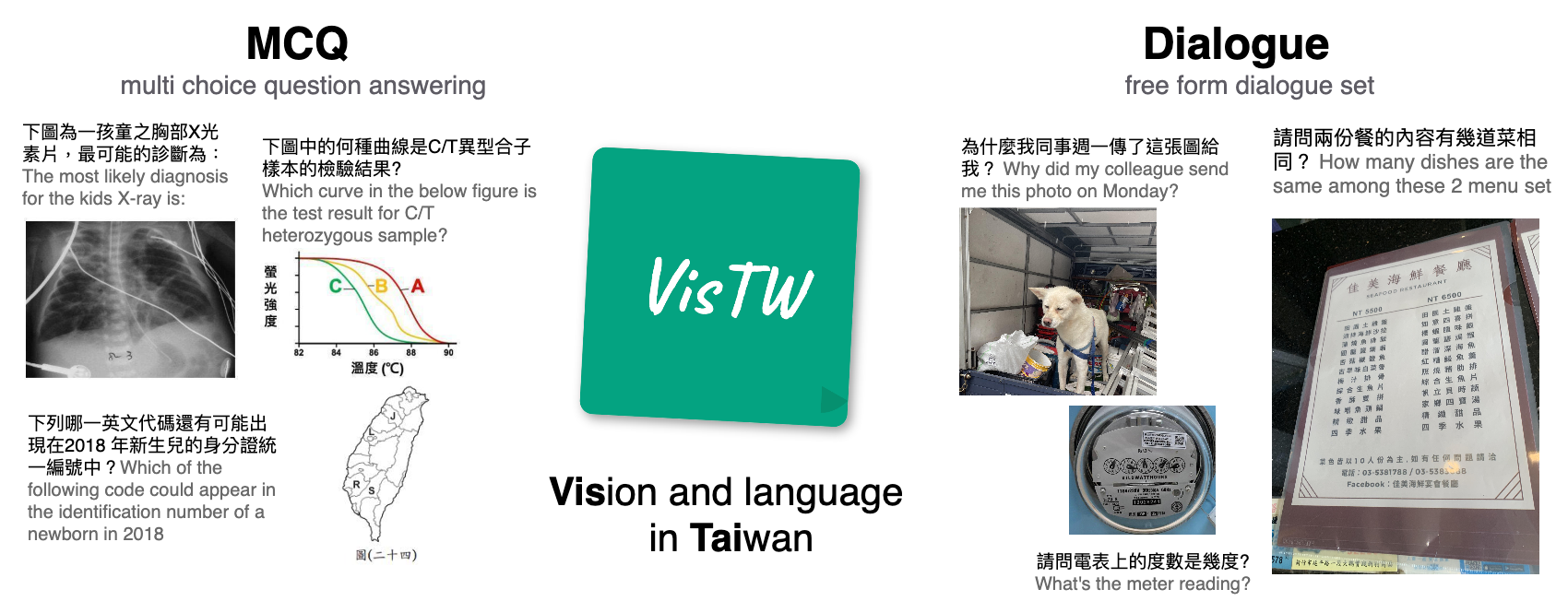}
    \caption{VisTW consists of two subsets: (1) MCQ - a collection of multiple-choice questions from 21 academic subjects (answer choices omitted for space); and (2) Dialogue - real-life images with corresponding questions requiring understanding of Traditional Chinese and Taiwan-specific cultural context.}
\label{fig:cover_introduction}
\end{figure}

Evaluating VLMs in Traditional Chinese presents unique challenges. Traditional Chinese characters are more complex and often do not map directly to their Simplified counterparts, causing potential inaccuracies when models trained on Simplified scripts encounter Traditional texts \citep{chen2024evlm}. Moreover, images from regions like Taiwan frequently incorporate culturally specific content and Traditional Chinese textual elements—such as signs, labels, or documents—introducing further contextual challenges not typically addressed by Western-centric or Simplified-Chinese-oriented datasets. Figure \ref{fig:cover_introduction} shows that 
Thus, dedicated benchmarks capturing both linguistic complexity and cultural nuances are essential for accurate evaluation.

Figure \ref{fig:cover_introduction} shows examples from our benchmark that illustrate the distinct challenges VLMs face when processing Taiwanese content. The left side demonstrates MCQ-style evaluations including geographic information correlated to specific to Taiwan-specific identification. The right side showcases dialogue scenarios requiring understanding of culturally-specific contexts such as deciphering Traditional Chinese menus and utility meters. These examples highlight content that cannot be adequately evaluated through simple character mapping from Simplified to Traditional Chinese, as they incorporate Taiwan-specific terminology, formats, and cultural knowledge.

\subsection{Our Contributions}

To fill this gap, we introduce two complementary benchmarks for evaluating VLMs in Traditional Chinese and Taiwanese contexts:

\paragraph{VisTW-MCQ.} This benchmark centers on structured, exam-style questions, reflecting high-level reasoning and subject-matter knowledge. Inspired by Taiwanese educational assessments, VisTW-MCQ features images (e.g., diagrams, charts) coupled with questions in Traditional Chinese. These questions often require interpretative or logical reasoning, offering a rigorous test bed for multimodal models beyond simple image-description tasks.

\paragraph{VisTW-Dialogue.} In contrast, VisTW-Dialogue targets open-ended scenarios rooted in everyday Taiwanese contexts. It includes image-question pairs on topics such as local landmarks, cultural practices, and daily life scenes containing Traditional Chinese text (e.g., street signs, menus). Rather than multiple-choice prompts, VisTW-Dialogue uses free-form queries, requiring models to produce answers in Traditional Chinese and handle unconstrained, real-world queries.

In summary, our contributions include: 
\begin{enumerate} 
    \item The first comprehensive vision-language benchmark for Traditional Chinese, featuring 3,795 test questions across 21 subjects (VisTW-MCQ) and 131 culturally-grounded dialogue pairs (VisTW-Dialogue).
    \item A validated, cost-effective VLM-as-judge evaluation framework with high human correlation ($\rho = 0.8466$) and a calibration approach ensuring evaluation continuity across model generations.
    \item Empirical evidence revealing performance gaps between Simplified and Traditional Chinese VLMs, with insights on model scaling efficiency and image resolution sensitivity across 31 different models.
\end{enumerate}
\section{Related Work}
\subsection{English-based Multimodal Benchmarks}
Vision-language benchmarks like ChartQA \citep{masry2022chartqa}, TextVQA \citep{singh2019towards}, and DocVQA \citep{mathew2021docvqa} have advanced visual text understanding through specialized tasks requiring OCR and reasoning. More recent exam-based evaluations such as MMMU \citep{yue2024mmmu} test expert-level multimodal reasoning across 30 subjects, pushing beyond surface-level image understanding to complex problem-solving. For open-ended generation, benchmarks like Vibe-Eval \citep{padlewski2024vibe} assess multimodal chat models with 269 visually-grounded prompts. However, these benchmarks remain English-centric with limited applicability to other languages and cultural contexts.
\subsection{Traditional Chinese and Multilingual Benchmarks}
For Traditional Chinese, recent benchmarks focus primarily on text-only evaluation. TMMLU \citep{hsu2023advancing} provides 3,300 multiple-choice questions from Taiwanese exams, while TMMLU+ \citep{tam2024tmmlu} expands this to 22,690 questions across 66 subjects. However, both lack multimodal components essential for vision-language evaluation. TMMBench included in the release of Breeze2 \citep{hsu2025breeze} is a multichoice traditional Chinese vision question that covers topics related to Taiwan, such as Taiwanese attractions, daily life, and Taiwan’s university entrance exams. However, as of this writing, the details and TMMBench dataset have not been publicly released yet.

Multilingual efforts include JMMU \citep{onohara2024jmmmu} for Japanese (though with only 30 questions per subjects) and CMMU~\citep{he2024cmmu} for Simplified Chinese (with limited public access and insufficient cultural context). M3Exam~\citep{zhang2023m3exam} covers 9 languages, including Simplified Chinese, but uses generic curricula without Traditional Chinese script or Taiwan-specific content.

This review reveals a critical gap: There is no comprehensive benchmark for evaluating vision-language models in Traditional Chinese that address both structured reasoning and open-ended generation while incorporating culturally relevant Taiwanese contexts. Our work on VisTW-MCQ and VisTW-Dialogue aims to address this specific need in the field.

\section{VisTW-MCQ : Visual Multi-Choice Question Answering}
\label{VisTW-MCQ}

The multi-choice question answering format is widely utilized to assess the knowledge and reasoning capabilities of Large Language Models (LLMs). In this section, we introduce a new visual multi-choice benchmark, \textbf{VisTW-MCQ}, composed of past examination questions from various educational levels in Taiwan, ranging from primary and secondary education to specialized undergraduate courses, such as veterinary medicine.

Our benchmark dataset was constructed using real-world exam papers collected from publicly available sources spanning the years 2013 to 2024. We selected subjects specifically requiring visual comprehension, such as medical diagnostics (e.g., interpreting X-ray and ultrasound images), geometry, electronic circuit design, and chemistry. 
The curation process involved manual extraction of questions and images from official exam archives and online PDFs. Each pair of questions and images was independently reviewed by a second annotator to verify content accuracy, clarity, and completeness, ensuring the absence of contextual errors, missing information, or invalid answer choices. To maintain consistent baseline guessing performance, the answer choices for each question were randomly shuffled, ensuring a uniform probability (approximately 25\%) of selecting the correct answer by chance.

In total, \textbf{VisTW-MCQ} includes 21 distinct subjects covering diverse areas such as medical diagnosis, nautical science with cartographic analysis, and technical disciplines. Each subject contains a minimum of 100 test questions, along with 5 few-shot learning examples as a development set, and 10 questions dedicated to validation purposes. For more details on image resolutions and aspect ratio we include this in Appendix \ref{app:resolution_mcq}. Table \ref{tab:showcase_different_dataset} compares \textbf{VisTW-MCQ} with existing visual multi-choice benchmarks, such as CMMU \citep{he2024cmmu}, JMMMU \citep{onohara2024jmmmu}, and ALM-Bench \citep{vayani2024alm}. Our dataset stands out due to its robustness and broader subject diversity.

\begin{table}[t!]
\centering
\small
\label{tab:benchmark_stats}
\begin{tabular}{lcrrr}
\toprule
\bf Dataset      & \bf Language & \bf Subjects & \bf Test~~ & \bf Validation \\
\midrule
CMMU~\citep{he2024cmmu}         & Simplified Chinese & 7              & 1,803       & 1,800\\
MMMU~\citep{yue2024mmmu} & English & 28  & 1,320   &   900\\
ALM-Bench~\citep{vayani2024alm}    & Traditional Chinese & 13 &  52 & 0\\
VisTW-MCQ  & Traditional Chinese & 21             & \bf 3,795       & 869\\
\bottomrule
\end{tabular}
\caption{Comparison of VisTW-MCQ with other similar datasets, for ALM-Bench, we only compare the subset for Traditional Chinese. }
\label{tab:showcase_different_dataset}
\end{table}

\subsection{Evaluation Method}

To evaluate the visual reasoning capabilities of Vision-Language Models (VLMs) across different subjects, we employ a zero-shot Chain-of-Thought prompting \citep{kojima2022large}. Each VLM is instructed to provide reasoning steps followed by the final answer choice explicitly. However, we observed that certain VLMs occasionally struggled to comply with formatting instructions, resulting in invalid responses. To mitigate evaluation bias due to parsing errors, we subsequently employed an auxiliary LLM parser ( gpt-4o-mini ) to extract and validate the final answer choice from each VLM response. We report the average scores for all 21 subjects as our final score.

However, one limitation of MCQA form of benchmark is that it does not reflect the real world use of VLM which mostly in free form question answering format. Hence we'll address this limitation by introducing another benchmark dataset.

\section{VisTW-Dialogue: Visual Free Form Dialogue Benchmark}

In this section, we introduce a visual free-form generation benchmark designed to bridge the gap between real-world user interactions and typical model evaluation procedures. Specifically, our goal is design a dataset which reflects real user experiences when interacting with VLMs in Traditional Chinese, where users naturally engage in open-ended dialogues rather than structured question-answering formats.

We manually curated both images and the corresponding dialogue prompts based on real-life scenarios encountered in Taiwan, such as interpreting metro rail maps or calculating shared bills from restaurant menus. Each image was intentionally captured by our researchers to ensure that these image has never been seen during VLM training. The researchers who took the photographs also tasked to write challenging questions related to the images, specifically designed to test spatial reasoning skills and local cultural knowledge unique to Taiwan.

To facilitate efficient data collection across diverse regions, we created a Discord server integrated with a data collection bot. We chose Discord as our data collection platform for several practical reasons: Participants already had Discord installed, it allowed direct photo uploads from mobile devices, and its threaded conversation structure facilitated discussion. Contributors, primarily existing Discord users, only needed to join the dedicated server to participate. The custom bot we developed served multiple functions: It notified users when data was received successfully, monitored emoji annotations for quality control, and provided a testbed for experimenting with various rating prompts within submission threads. Additional details of the labeling can be found in the Appendix \ref{app:discord_labeling}.

Following the initial submission, our quality control process involved multiple stages:

\textbf{1. Answer Provision:} The researcher who submitted the image was required to provide an answer to their question within a newly created submission thread.

\textbf{2. Peer Review:} Other team members reviewed the provided answers, making corrections or improvements when necessary.

\textbf{3. Community Moderation:} We implemented a simple but effective quality control mechanism using Discord's emoji reactions. Members could flag inappropriate or unsuitable submissions using downvotes, which were automatically recorded by our bot.

\textbf{4. Filtering Process:} Submissions that received more than two downvotes (typically from a user and a moderator) were removed during the dataset cleaning phase.

The data collection phase spanned from July 2024 to February 2025, resulting in a dataset comprising 131 unique image-question pairs with ground truth. The image resolutions and aspect ratio of VisTW-Dialogue are included in Appendix \ref{app:resolution_dialogue}.

\subsection{Automatic Evaluation : VLM as a Judge}

To reduce the barrier of quick evaluations, we used a VLM as an automated judge to score responses based on questions, images, and ground truth answers. Using LLMs to evaluate free-form generation has become standard practice, exemplified by benchmarks such as AlpacaEval \citep{alpaca_eval} and MT-Bench \citep{zheng2023judging}.

Inspired by Vibe-Eval \citep{padlewski2024vibe}, we crafted an evaluation prompt that explicitly includes scoring criteria, the original question, the assistant's response, the ground truth and optionally, the image itself. 
The LLM judge is instructed to first provide a detailed explanation followed by the final numerical score from 0 to 10, the detail breakdown of score criteria can be found in Appendix \ref{app:dialogue_judge_prompt}.

To validate the reliability of LLM-generated scores, we conducted human evaluations across four selected models (gpt-4o, claude-3-5-sonnet, gpt-4o-mini, claude-3-7-sonnet). Each response was independently scored by three human annotators, and the average rating was caculated to establish a human evaluation baseline. We evaluated several prominent VLMs including gemini-2.0-pro, gemini-2.0-flash, Qwen 2.5 VL 72B, claude-sonnet-3-5-20241022, gpt-4o-2024-08-06, LLaMA Vision 90B, and Qwen VL 72B. Table \ref{tab:eval_correlations} reports the Spearman correlations between LLM-generated scores and human evaluations. 

Our analysis revealed that incorporating images in the evaluation prompts generally improved model-human score correlations, contradicting earlier findings by \citet{padlewski2024vibe}, which indicated minimal improvements from multimodal judging. Notably, the gemini-2.0-flash-001 model achieved performance comparable to claude-3-5-sonnet-20241022, despite being approximately 36 times less costly. When we averaged the scores from all VLM judges into an ensemble approach, we found it exhibited the strongest correlation with human scores, reaching 0.8466 without images and 0.8463 with images, suggesting that combining diverse model judgments provides more robust evaluation than any single model alone.

All reported correlations were statistically significant with $p < 0.001$. The strongest correlation was observed with claude-3-5-sonnet-20241022 ($\rho = 0.8336$, $p < 0.001$) when images were included. Given these results and future cost considerations, we selected gemini-2.0-flash-001 as our primary LLM judge. The complete evaluation of the performance of one model on VisTW-Dialogue cost merely \$0.05 USD when using gemini-2.0-flash-001 as the automated judge.

Additionally, we found no significant correlation between response length and automated judge scores with a Spearman's $\rho = 0.26789$, $p < 0.001$, indicating that our evaluation framework does not exhibit substantial length bias. Our human annotated data are available public for calibrate newer VLMs against human preference. Additional details can be found in Appendix \ref{app:dialogue_judge_prompt}.

\begin{table}[t]
\begin{center}
\begin{tabular}{lcc}
\toprule
\textbf{VLM Judge} & \textbf{With Image} & \textbf{Without Image} \\
\midrule
claude-3-5-sonnet-20241022 & \textbf{0.8336} & 0.8282 \\
gemini-2.0-flash-001 & \textbf{0.8241} & 0.8194 \\
gpt-4o-2024-08-06 & 0.7985 & \textbf{0.8277} \\
gemini-2.0-pro-exp-02-05 & 0.8196  & \textbf{0.8320} \\
qwen2.5-vl-72b-instruct & \textbf{0.7501} & 0.7341 \\
gpt-4o-mini-2024-07-18 & 0.7349 & \textbf{0.7360} \\
\midrule
Ensemble all models & 0.8463 & \textbf{0.8466} \\
\bottomrule
\end{tabular}
\caption{Spearman correlations between human-assigned scores and scores generated by different VLMs serving as automated judges. \textit{With Image} indicates that the evaluation prompt included the associated image as additional context, whereas \textit{Without Image} denotes that only the textual question and ground-truth answer were provided to the judging model. Higher correlations indicate better alignment with human judgment.}
\label{tab:eval_correlations}
\end{center}
\end{table}

\section{Benchmark Results}

We evaluated $X$ vision-language models (VLMs) spanning both closed and open-weight categories as presented in Table~\ref{tab:overall_result}. Detailed model specifications are provided in Appendix~\ref{app:model_details}. Our analysis reveals strong correlation between VisTW-Dialogue and VisTW-MCQ performance metrics (Spearman's $\rho = 0.7746$, $p < 0.001$; Kendall's $\tau = 0.6086$, $p < 0.001$).

The two Traditional Chinese VLMs in our evaluation, Breeze2-3B and Breeze2-8B, demonstrate inferior performance compared to Simplified Chinese counterparts of similar parameter scales, such as InternVL2.5-4B and InternVL2-8B variants. An intriguing observation emerges with Gemini-2.0-flash-thinking, which achieves exceptionally high scores in VisTW-Dialogue (6.51) but relatively modest performance in VisTW-MCQ (0.376). Conversely, deepseek-ai-deepseek-vl2-small scores 0.318 in VisTW-MCQ while nearly failing in VisTW-Dialogue. Upon manual validation, we discovered this model predominantly outputs bounding boxes for object detection tasks while neglecting other user requirements in VisTW-Dialogue. Interestingly, its smaller counterpart (tiny version) manages to partially address VisTW-Dialogue questions, suggesting potential training anomalies in the small variant.

Examining the performance of smaller VLMs reveals interesting efficiency-capability tradeoffs. Models with fewer parameters ($\leq$4B) show consistent degradation in VisTW-Dialogue scores, with OpenGVLab-InternVL2-1B scoring just 2.13 compared to its 8B counterpart's 3.45. However, this performance gap narrows considerably for VisTW-MCQ, where InternVL2-1B achieves 0.2689 versus InternVL2-8B's 0.3431---only a 21.6\% reduction despite using 87.5\% fewer parameters. This suggests smaller models maintain reasonable factual reasoning capabilities while struggling more with open-ended, multimodal tasks. Notably, the InternVL2 series demonstrates graceful performance scaling, with consistent improvements as parameter count increases from 1B to 8B in both metrics, making them promising candidates for resource-constrained applications.

\begin{table}[t!]
\centering
\begin{tabular}{lccccc}
\toprule
\multirow{2}{*}{\textbf{Model}} & \multicolumn{2}{c}{\textbf{VisTW-Dialogue}} & \multicolumn{2}{c}{\textbf{VisTW-MCQ}} & \textbf{Avg} \\
\cmidrule(lr){2-3} \cmidrule(lr){4-5}
 & \textbf{Score 0-10} & \textbf{Rank} & \textbf{Accuracy} & \textbf{Rank} & Rank \\
\midrule
gemini-2.0-pro-exp-02-05 & 6.72 & 1 & 0.6619 & 1 & 1.0 \\
gemini-2.0-flash-001 & 6.15 & 3 & 0.6596 & 2 & 2.5 \\
gpt-4o-2024-11-20 & 6.12 & 4 & 0.5755 & 4 & 4.0 \\
claude-3-5-sonnet-20241022 & 5.96 & 6 & 0.6019 & 3 & 4.5 \\
gemini-2.0-flash-lite-preview-02-05 & 5.92 & 7 & 0.4992 & 6 & 6.5 \\
Qwen2.5-VL-72B-instruct & 4.87 & 9 & 0.5413 & 5 & 7.0 \\
gemini-2.0-flash-thinking-exp-1219 & 6.51 & 2 & 0.3764 & 14 & 8.0 \\
gemini-1.5-pro & 5.05 & 8 & 0.4417 & 8 & 8.0 \\
gpt-4o-2024-08-06 & 5.98 & 5 & 0.4000 & 12 & 8.5 \\
Qwen2-VL-72B-instruct & 4.21 & 13 & 0.4701 & 7 & 10.0 \\
gpt-4o-mini-2024-07-18 & 4.74 & 10 & 0.4091 & 11 & 10.5 \\
gemini-1.5-flash & 4.26 & 12 & 0.3943 & 13 & 12.5 \\
gemma3-27b-it & 3.94 & 16 & 0.4375 & 9 & 12.5 \\
Qwen2.5-VL-7B-Instruct & 4.54 & 11 & 0.3592 & 15 & 13.0 \\
Llama-3.2-90B-Vision-Instruct & 3.44 & 22 & 0.4119 & 10 & 16.0 \\
InternVL2.5-8B & 3.90 & 17 & 0.3447 & 17 & 17.0 \\
gemini-1.5-flash-8B & 4.18 & 15 & 0.3280 & 22 & 18.5 \\
claude-3-haiku-20240307 & 3.70 & 18 & 0.3291 & 20 & 19.0 \\
InternVL2-8B & 3.45 & 21 & 0.3431 & 18 & 19.5 \\
Qwen2-VL-7B-Instruct & 4.21 & 13 & 0.3004 & 26 & 19.5 \\
InternVL2.5-4B & 3.60 & 20 & 0.3291 & 21 & 20.5 \\
nova-lite-v1 & 3.26 & 23 & 0.3376 & 19 & 21.0 \\
Llama-3.2-11B-Vision-Instruct & 2.58 & 27 & 0.3262 & 23 & 25.0 \\
Breeze2-8B-Instruct & 3.14 & 24 & 0.2915 & 28 & 26.0 \\
Breeze2-3B-Instruct & 2.90 & 26 & 0.2971 & 27 & 26.5 \\
InternVL2-4B & 2.31 & 28 & 0.3081 & 25 & 26.5 \\
deepseek-vl2-small & 0.51 & 31 & 0.3181 & 24 & 27.5 \\
CogVLM2-llama3-chinese-chat & 2.96 & 25 & 0.2777 & 31 & 28.0 \\
InternVL2-2B & 2.22 & 29 & 0.2891 & 29 & 29.0 \\
deepseek-vl2-tiny & 2.01 & 30 & 0.2781 & 30 & 30.0 \\
\bottomrule
\end{tabular}
\caption{For VisTW-MCQ we prompt our model in zero shot chain of thought fashion and report the average score from all 23 subjects. Models are sorted by average rank from both subsets.}
\label{tab:overall_result}
\end{table}
\section{Analysis}

\subsection{Sensitivity of Image Resolution}

We studied how image resolution affects model performance through systematic experiments. For VisTW-MCQ, we tested both directions: upscaling to $2\times$ and $4\times$ the original resolution using the state-of-the-art SwinIR model \citep{liang2021swinir}, as well as downscaling to $1/2$, $1/4$, and $1/8$ of the original size. For VisTW-Dialogue, due to the initially large image sizes in this dataset, we only conducted downscaling experiments at the scales $1/2$, $1/4$, and $1/8$.

\begin{figure}
    \centering
    \includegraphics[width=1.0\columnwidth]{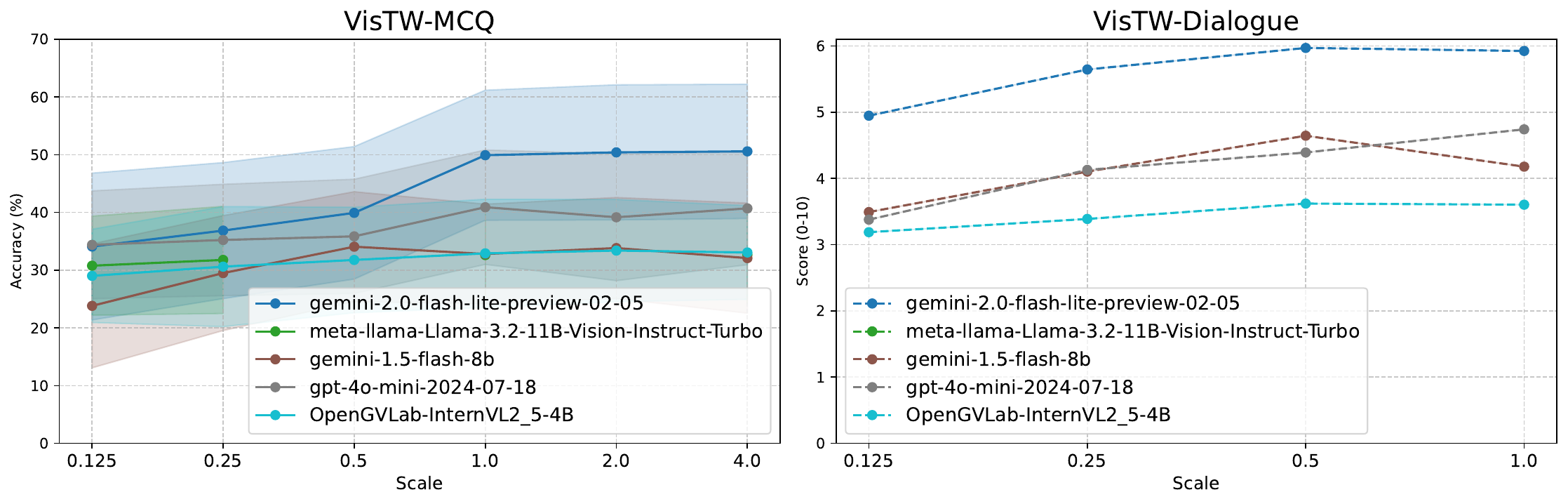}
    \caption{Effect of image resolution scaling on model performance. Left: Performance on VisTW-MCQ with both upscaled ($2\times$, $4\times$) and downscaled ($1/2$, $1/4$, $1/8$) images. Right: Performance on VisTW-Dialogue with downscaled images.}
    \label{fig:scale_over_scores_in_exam_free}
\end{figure}

Figure~\ref{fig:scale_over_scores_in_exam_free} illustrates our findings on resolution sensitivity. For VisTW-MCQ, performance remained consistent across resolution variations, suggesting that even lower resolutions contained sufficient information for multiple-choice tasks. In contrast, for VisTW-Dialogue, model performance stabilized at approximately half the original resolution. We observed model-specific variations: gemini-2.0-flash-lite showed slight improvements with increased resolution, while gemini-1.5-flash-8B exhibited minor performance declines, indicating different architectural sensitivities to input resolution.

\subsection{Comparison between With and Without Vision Cues}
In the previous section, we showed score degradation with reduced image resolution. Here, we examine performance when image context is entirely omitted. For MCQ tasks, this analysis establishes a baseline that measures how much linguistic knowledge alone contributes to task performance before visual information becomes beneficial. Similarly, for dialogue tasks, images-free score serve as baseline measurements, indicating how significantly performance drops when visual context is absent. We conducted this experiment using the same models evaluated in the previous section. Table \ref{tab:results_without_image} presents the results, which show consistent score reductions across all four models in both datasets when images are withheld. The performance degradation is expected, as it indicates that VLMs struggle when visual information is not provided.

\begin{table}[t]
\begin{center}
\begin{tabular}{lcccc}
\toprule
\multirow{2}{*}{\textbf{Model}} & \multicolumn{2}{c}{\textbf{VisTW-MCQ}} & \multicolumn{2}{c}{\textbf{VisTW-Dialogue}} \\
\cmidrule(lr){2-3} \cmidrule(lr){4-5}
 & \textbf{w/ img} & \textbf{w/o img} & \textbf{w/ img} & \textbf{w/o img} \\
\midrule
gemini-2.0-flash-lite-preview-02-05 & 0.50 & 0.32 & 5.92 & 2.35 \\
gpt-4o-mini-2024-07-18 & 0.41 & 0.33 & 4.74 & 1.95 \\
gemini-1.5-flash-8b & 0.33 & 0.26 & 4.18 & 1.85 \\
OpenGVLab-InternVL2\_5-4B & 0.33 & 0.29 & 3.60 & 1.29 \\
\bottomrule
\end{tabular}
\caption{Performance comparison of various multimodal models across different evaluation tasks. MCQ scores represent accuracy, while dialogue scores represent quality ratings. Higher scores indicate better performance.}
\label{tab:results_without_image}
\end{center}
\end{table}

\subsection{Correlations with Other Benchmarks}
\begin{figure}
    \centering
    \includegraphics[width=1.0\columnwidth]{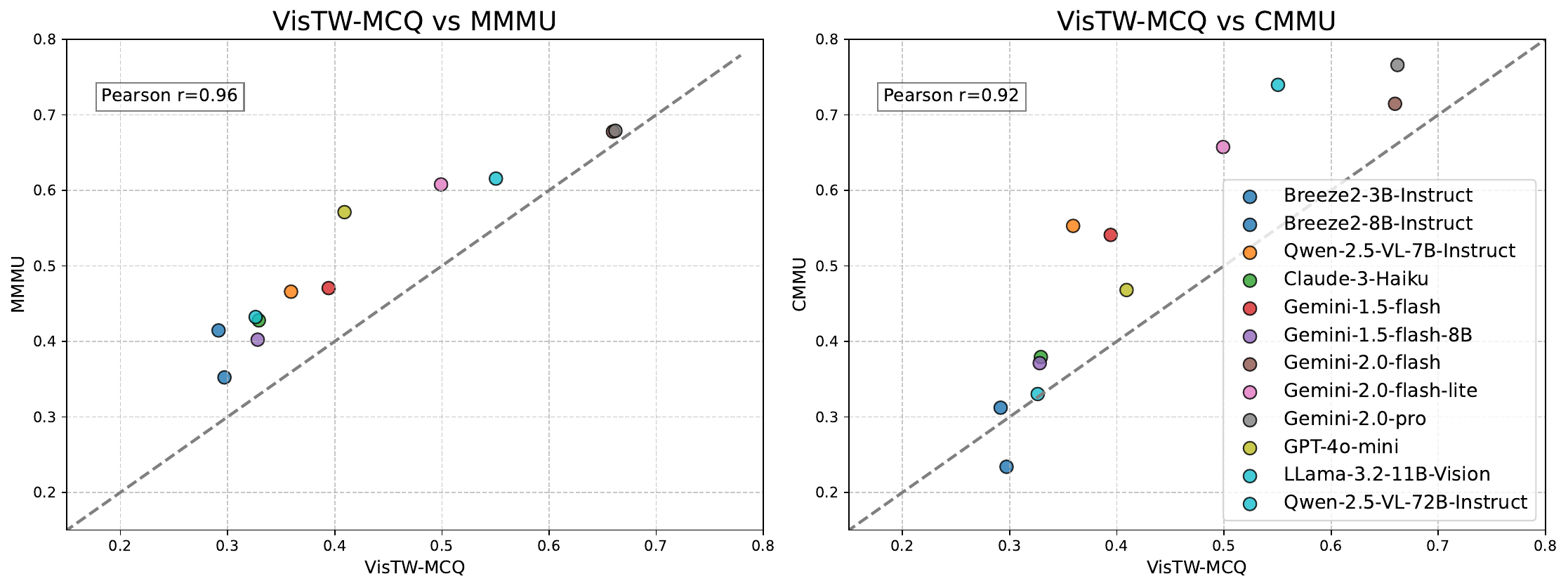}
    \caption{Comparison of VisTW-MCQ scores with MMMU (left) and CMMU (right) on a selected subset of models of varying scales. We observe a rough correlation across the three benchmarks, though some deviations suggest differences in the specific knowledge or reasoning skills each test emphasizes.}
    \label{fig:correlation_bench}
\end{figure}
As shown in Figure \ref{fig:correlation_bench}, we compare our VisTW-MCQ results with exam style benchmarks: MMMU~\citep{yue2024mmmu} and CMMU~\citep{he2024cmmu} on a selected set of models spanning different scales. Despite the relatively small subset (due to computational and financial constraints), we observe a broadly consistent trend: models that perform well on MMMU or CMMU also tend to score higher on VisTW-MCQ. However, we also note that some points deviate from the main diagonal, suggesting that the three benchmarks are not fully interchangeable. The models show stronger relative performance on MMMU and CMMMU compare to VisTW-MCQ scores.

\subsection{Contingency Plan When the Gemini-2.0-Flash Is Retired}
Due to our choice of using a close weight VLM : Gemini 2.0 Flash as our judge; this means it will eventually become unavailable when newer versions become available. Consequently, scores assigned by a retired model, such as \emph{gemini-2.0-flash} will become difficult to compare with those of the newer VLMs. Many LLM-as-judge works overlook this scenario, typically re-running an entire benchmark using the new model as judge. However, large-scale evaluations make such an approach infeasible.

\begin{figure}
    \centering
    \includegraphics[width=0.9\columnwidth]{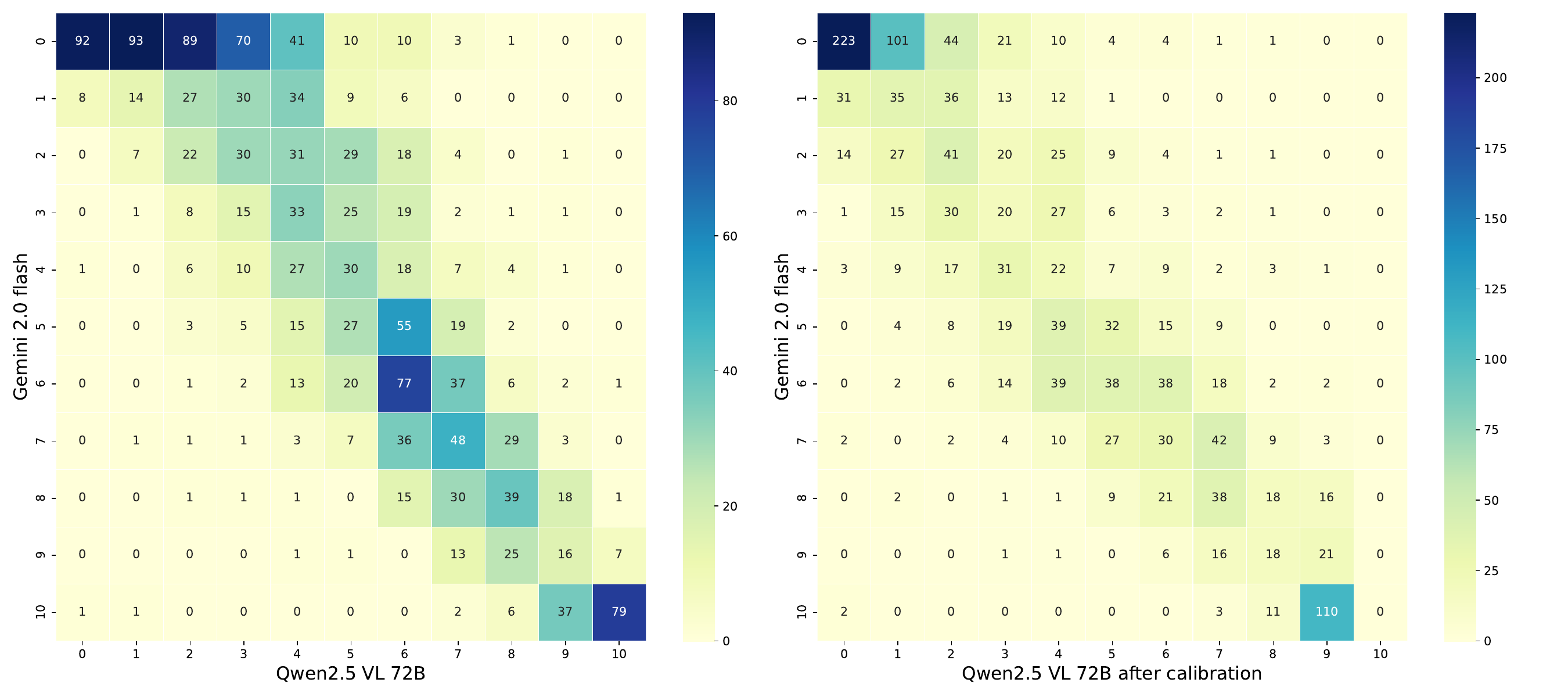}
    \caption{Comparison of Gemini 2.0 flash and Qwen 2.5 VL 72B score distribution rounded to the nearest integer (left) and Qwen 2.5 VL 72B scores after calibrated against Gemini 2.0 flash (right). }
    \label{fig:vlm_judge_correlation_distribution}
\end{figure}

To address this, we propose a calibration-based contingency plan using \emph{Qwen2.5-VL~72B}, the highest performing VLM in Table \ref{tab:overall_result}. Figure~\ref{fig:vlm_judge_correlation_distribution} (left) shows that the raw scores for Qwen2.5-VL are systematically higher than those from gemini-2.0-flash. We therefore fit a linear mapping (based on five-vote distributions) to align Qwen2.5-VL scores with gemini-2.0-flash. After calibration, instances that gemini-2.0-flash scores as zero now cluster near zero under Qwen2.5-VL, however the number of perfect scores is reduced (Figure~\ref{fig:vlm_judge_correlation_distribution}, right).  After calibration, the spearman correlation with human score and ensemble of Qwen2.5-VL scores improve from 0.7993 to 0.8070.

Table~\ref{tab:judge_scores} compares raw vs.\ calibrated Qwen2.5-VL scores to gemini-2.0-flash on \emph{VisTW-Dialogue}, demonstrating how calibration improves alignment. This result offers a practical strategy to maintain continuity of evaluation even if the original judge model is no longer available.

\begin{table}[t]
\begin{center}
\begin{tabular}{lcccc}
\toprule
\textbf{Model} & \textbf{Gemini} & \textbf{Qwen} & \textbf{Qwen (calibrated)} & \textbf{$\Delta$ (Cal - Gem)} \\
\midrule
gemini-2.0-flash-001 & 6.15 & 6.65 & 5.73 & -0.42 \\
qwen2.5-vl-72b-instruct & 4.87 & 6.18 & 5.12 & +0.25 \\
Llama-Breeze2-8B-Instruct & 3.14 & 4.12 & 3.16 & +0.02 \\
Llama-Breeze2-3B-Instruct & 2.90 & 4.07 & 3.06 & +0.16 \\
\bottomrule
\end{tabular}
\caption{Comparison of scores assigned by different judge models (Gemini, Qwen, and calibrated Qwen) across various evaluated models. The $\Delta$ column shows the difference between calibrated Qwen VL scores and Gemini scores, highlighting the effectiveness of the calibration. Values close to zero indicate better alignment.}
\label{tab:judge_scores}
\end{center}
\end{table}

\subsection{Direct Answer vs CoT in VisTW-MCQ}
To analyze the impact of reasoning on performance, we compare two prompting strategies: Zero-Shot Chain of Thought (CoT) and Direct Answer prompting. The latter asks models to provide answers without intermediate reasoning steps. Table~\ref{tab:cot_vs_da_perf} presents the comparative results.

Interestingly, CoT does not universally benefit all vision-language models (VLMs). The performance gain from reasoning is predominantly observed in models that already demonstrate strong direct answer capabilities. For instance, Gemini-2.0-Flash-001, despite being smaller than Gemini-2.0-Pro (as evidenced by its lower direct answer performance), exhibits a substantially larger improvement margin (13.9\%) when employing reasoning steps.

Our analysis reveals that high-performing models generally gain 2-10\% improvement through CoT prompting. Conversely, lower-performing VLMs show degraded performance with CoT, as evidenced by the negative differences in the lower portions of Table~\ref{tab:cot_vs_da_perf}. We hypothesize that inferior vision processing capabilities in these models may introduce hallucinations when given extended reasoning space, ultimately diminishing their performance.

\begin{table}[htbp]
\centering
\begin{tabular}{lccc}
\toprule
\textbf{Model Name} & \textbf{CoT} & \textbf{Direct Answer} & \textbf{Difference} \\
\midrule
gemini-2.0-flash-001 & 0.660 & 0.521 & 0.139 \\
gemini-2.0-pro-exp-02-05 & 0.662 & 0.569 & 0.093 \\
gpt-4o-2024-11-20 & 0.576 & 0.486 & 0.090 \\
claude-3-5-sonnet-20241022 & 0.602 & 0.519 & 0.083 \\
qwen2.5-vl-72b-instruct & 0.502 & 0.433 & 0.069 \\
gpt-4o-mini-2024-07-18 & 0.409 & 0.350 & 0.059 \\
Llama-Breeze2-3B-Instruct & 0.297 & 0.267 & 0.030 \\
gemini-1.5-flash & 0.394 & 0.371 & 0.023 \\
gemini-1.5-pro & 0.442 & 0.420 & 0.022 \\
gemini-1.5-flash-8b & 0.328 & 0.309 & 0.019 \\
deepseek-ai-deepseek-vl2-tiny & 0.278 & 0.259 & 0.019 \\
OpenGVLab-InternVL2-8B-MPO & 0.353 & 0.338 & 0.015 \\
Qwen-Qwen2-VL-7B-Instruct & 0.300 & 0.287 & 0.013 \\
Qwen2.5-VL-7B-Instruct & 0.359 & 0.346 & 0.013 \\
Llama-3.2-90B-Vision-Instruct-Turbo & 0.412 & 0.404 & 0.008 \\
THUDM-cogvlm2-19B & 0.278 & 0.277 & 0.001 \\
claude-3-haiku-20240307 & 0.329 & 0.330 & -0.001 \\
OpenGVLab-InternVL2-8B & 0.343 & 0.347 & -0.004 \\
OpenGVLab-InternVL2-2B & 0.289 & 0.294 & -0.005 \\
OpenGVLab-InternVL2\_5-8B & 0.345 & 0.352 & -0.007 \\
OpenGVLab-InternVL2-4B & 0.308 & 0.315 & -0.007 \\
Llama-Breeze2-8B-Instruct & 0.292 & 0.299 & -0.007 \\
Llama-3.2-11B-Vision-Instruct & 0.326 & 0.335 & -0.009 \\
deepseek-ai-deepseek-vl2-small & 0.318 & 0.327 & -0.009 \\
OpenGVLab-InternVL2-1B & 0.269 & 0.282 & -0.013 \\
OpenGVLab-InternVL2\_5-4B & 0.329 & 0.345 & -0.016 \\
\bottomrule
\end{tabular}
\caption{Model Performance Comparison: CoT vs Direct Answer (Sorted by Difference)}
\label{tab:cot_vs_da_perf}
\end{table}

\section{Conclusion}
In this work, we introduce VisTW-MCQ and VisTW-Dialogue, the first comprehensive benchmarks for evaluating Vision-Language Models in Traditional Chinese. Our preliminary results reveal that VLMs specifically trained on Traditional Chinese still significantly underperform compared to their Simplified Chinese counterparts when processing visual content in Traditional Chinese contexts. Our results show a strong correlation between structured and free-form task performance suggests consistent capabilities across different evaluation formats while measuring different aspects of VLMs. Moreover, our analysis on factors such as image resolution and model scaling provides valuable insights into the practical deployment of these models. Future work should focus on improving the performance of the model on traditional Chinese content and expanding the benchmark to include a more diverse cultural context in Taiwan.

\section*{Ethics Statement}
All images in VisTai-Dialogue were collected by our researchers specifically for this benchmark, with contributors explicitly waiving ownership rights and consenting to public use of their images. We implemented strict guidelines to ensure no personally identifiable information (PII) was included—images containing names, faces of non-public figures, Wi-Fi passwords, or residential addresses were prohibited and filtered through community moderation. For VisTai-MCQ, we used only publicly available exam questions from official sources spanning 2013-2024, maintaining academic integrity while respecting intellectual property.

Our work directly addresses the underrepresentation of Traditional Chinese in vision-language benchmarks, promoting greater linguistic and cultural inclusivity in AI evaluation. By creating resources specifically for Traditional Chinese, we help ensure that NLP advancements benefit communities beyond those using dominant languages like English or Simplified Chinese. Our benchmark incorporates diverse geographical sampling across Taiwan through contributions from nine individuals in different regions, though we acknowledge a concentration of samples from Taipei.

\bibliography{vistai}
\bibliographystyle{colm2025_conference}

\appendix
\section{Appendix}
\section{Creating VisTW-Dialogue on Discord}
\label{app:discord_labeling}
\textbf{Guidelines:} The collected question-image pairs were designed to be challenging for Vision Language Models (VLMs), focusing on realistic scenarios where users might seek assistance (e.g., for visual impairment). All contributors were instructed to ensure their submissions contained no personally identifiable information (PII), including names or faces of non-public figures, Wi-Fi passwords, or any information that could reveal residential addresses.
The dataset comprises question-image pairs contributed by 9 different individuals from various locations across Taiwan. While the geographic distribution covers multiple regions, there is a higher concentration of samples from Taipei (the capital of Taiwan) as the majority of our contributors resided there.

\subsection{Human Annotation Process}
\label{app:discord_labeling_human_process}

\begin{figure}[h]
    \centering
    \includegraphics[width=\textwidth]{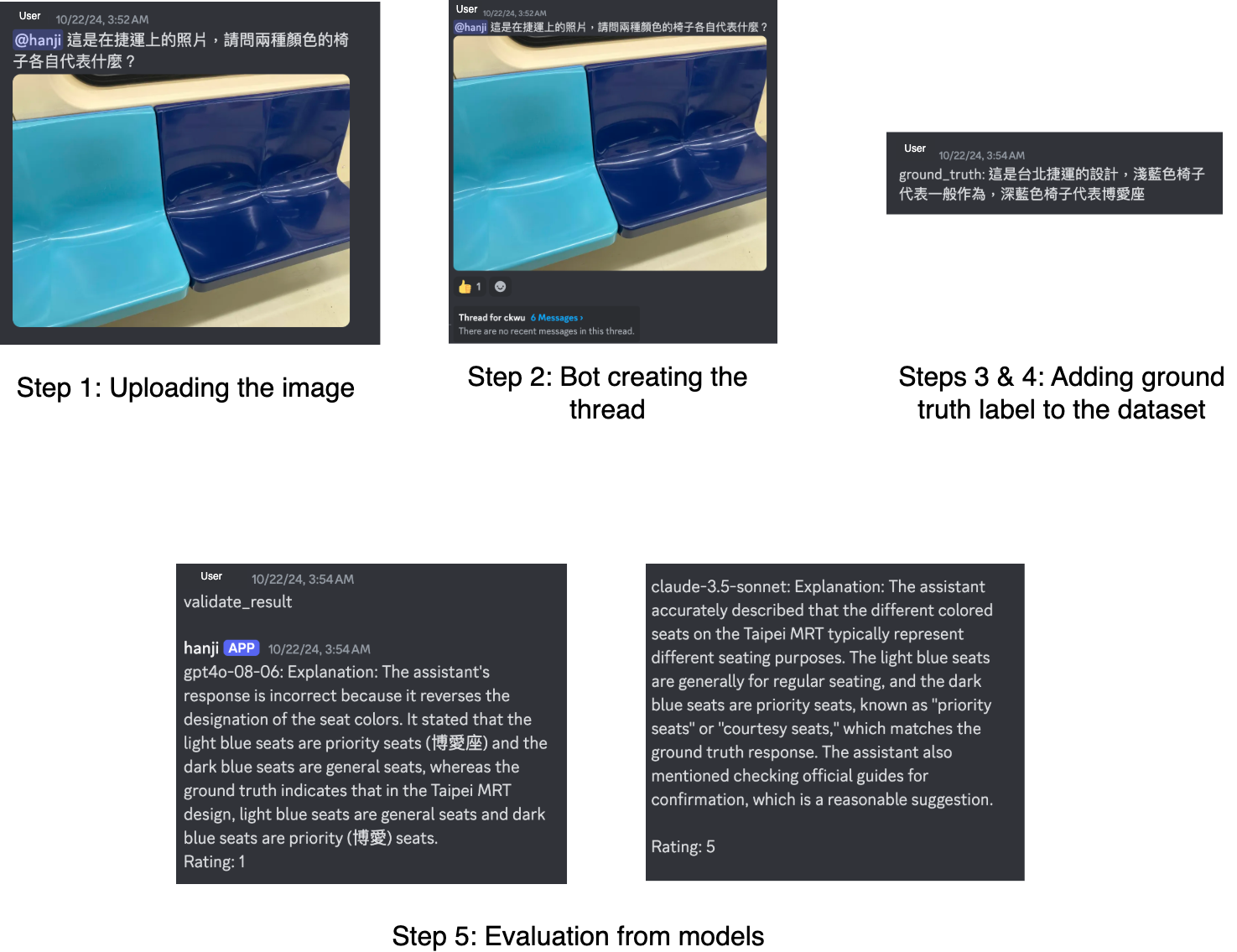}
    \caption{Workflow of adding images, labeling, and evaluating}
    \label{fig:workflow}
\end{figure}

As illustrated in Figure \ref{fig:workflow}, steps to label a problem are:
\begin{enumerate}
\item Upload an image, use @ + hanji to submit it to the bot on discord, and include the question and ground\_truth $<$standard answer$>$
\item The bot will create a discussion thread, or people can wait until the thread appears before adding the standard answer.
\item Format for updating the standard answer: as long as the text mentions "ground\_truth:", the system will automatically capture the text that follows as the standard answer.
\item The standard answer can be updated repeatedly using step 3.
\item People can type "validate\_result" to get score evaluations for responses from two models.
\end{enumerate}
Note: When people upload images, they waive ownership rights and agree to make the photos public.





\clearpage
\section{Additional Details for VisTW-MCQ and VisTW-Dialogue}

\subsection{Details of subjects for VisTW-MCQ}

Table \ref{tab:subject_overview} shows each of the subjects found in VisTW-MCQ with their broad category, as well as total number of questions.

\begin{CJK}{UTF8}{bsmi}
\begin{table}[t]
\begin{center}
\begin{tabular}{lcccc}
\toprule
\textbf{Subject Name} & \textbf{Chinese Name} & \textbf{Test} & \textbf{Val} & \textbf{Dev} \\
\midrule
Accounting & 會計學 & 100 & 29 & 5 \\
Arts & 藝術 & 385 & 91 & 5 \\
Biology & 生物學 & 150 & 8 & 5 \\
Chemistry & 化學 & 172 & 38 & 5 \\
Chinese Literature & 中國文學 & 100 & 45 & 5 \\
Dentistry & 牙醫學 & 220 & 49 & 5 \\
Electronic Circuits & 電子電路 & 388 & 91 & 5 \\
Fundamentals of Physical Therapy & 物理治療基礎 & 150 & 23 & 5 \\
Geography & 地理學 & 100 & 39 & 5 \\
Mathematics & 數學 & 240 & 54 & 5 \\
Mechanics & 力學 & 217 & 49 & 5 \\
Medical & 醫學 & 172 & 37 & 5 \\
Music & 音樂 & 100 & 6 & 5 \\
Natural Science & 自然科學 & 303 & 70 & 5 \\
Navigation & 航海學 & 100 & 16 & 5 \\
Pharmaceutical Chemistry & 藥物化學 & 100 & 45 & 5 \\
Physics & 物理學 & 100 & 23 & 5 \\
Sociology & 社會學 & 348 & 81 & 5 \\
Statistics & 統計學 & 100 & 45 & 5 \\
Structural Engineering & 結構工程 & 100 & 19 & 5 \\
Veterinary Medicine & 獸醫學 & 150 & 11 & 5 \\
\bottomrule
\end{tabular}
\caption{Overview of subjects with VisTW-MCQ splits}
\label{tab:subject_overview}
\end{center}
\end{table}
\end{CJK}

\subsection{VisTW-MCQ}
\label{app:resolution_mcq}

Figure \ref{fig:aspect_ratio_distribution} (left) shows the aspect ratio of mostly long rectangle, most of the width is longer than the height. In the Figure \ref{fig:resolution_distribution} (left) for the MCQ dataset most of the image pixel is below 1M pixels. 

\subsection{VisTW-Dialogue}
\label{app:resolution_dialogue}

\begin{figure}
    \centering
    \includegraphics[width=0.8\columnwidth]{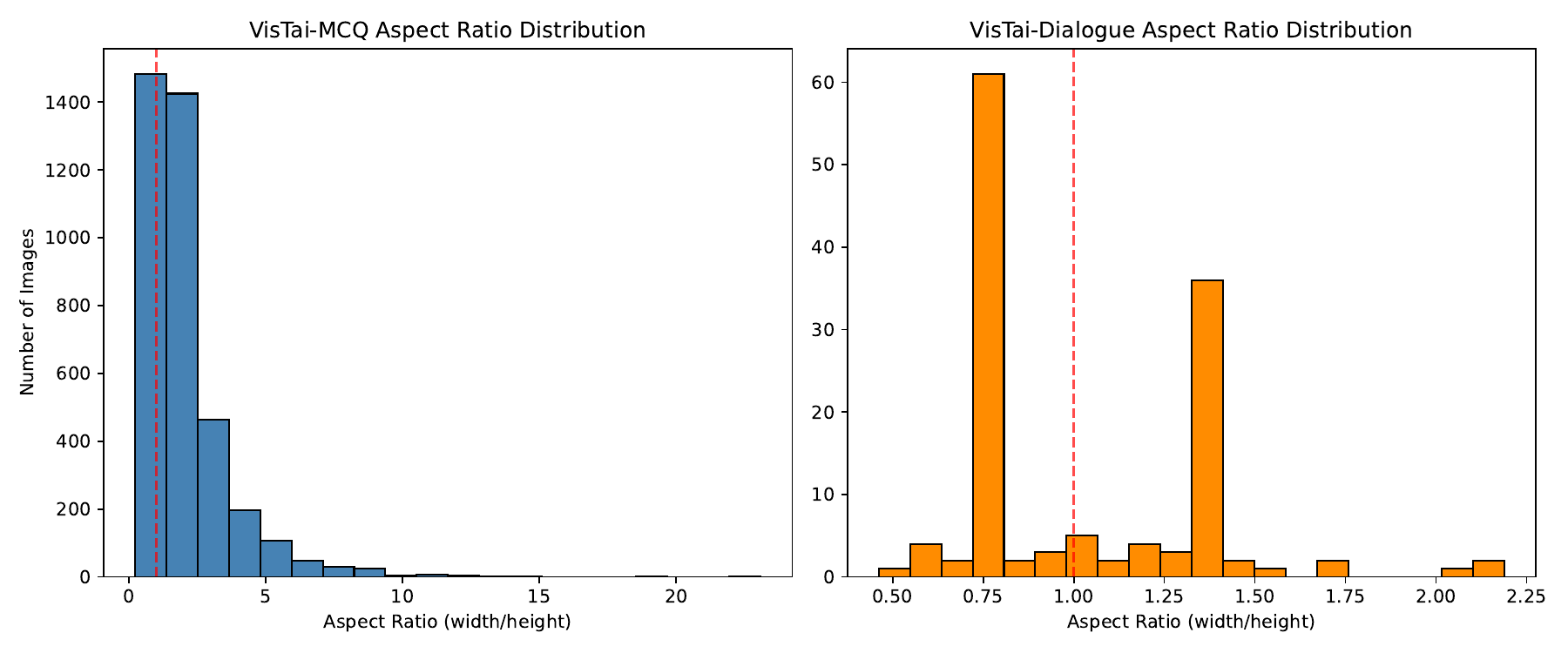}
    \caption{Histograms comparing the aspect ratio distributions (width/height) of the VisTW-MCQ dataset (left) and the VisTW-Dialogue dataset (right). The vertical dashed line on the right indicates an aspect ratio of 1 (i.e., square). Most MCQ images fall below an aspect ratio of 1 (portrait), while the Dialogue dataset shows a broader spread, with many images close to square.}
    \label{fig:aspect_ratio_distribution}
\end{figure}

\begin{figure}
    \centering
    \includegraphics[width=0.8\columnwidth]{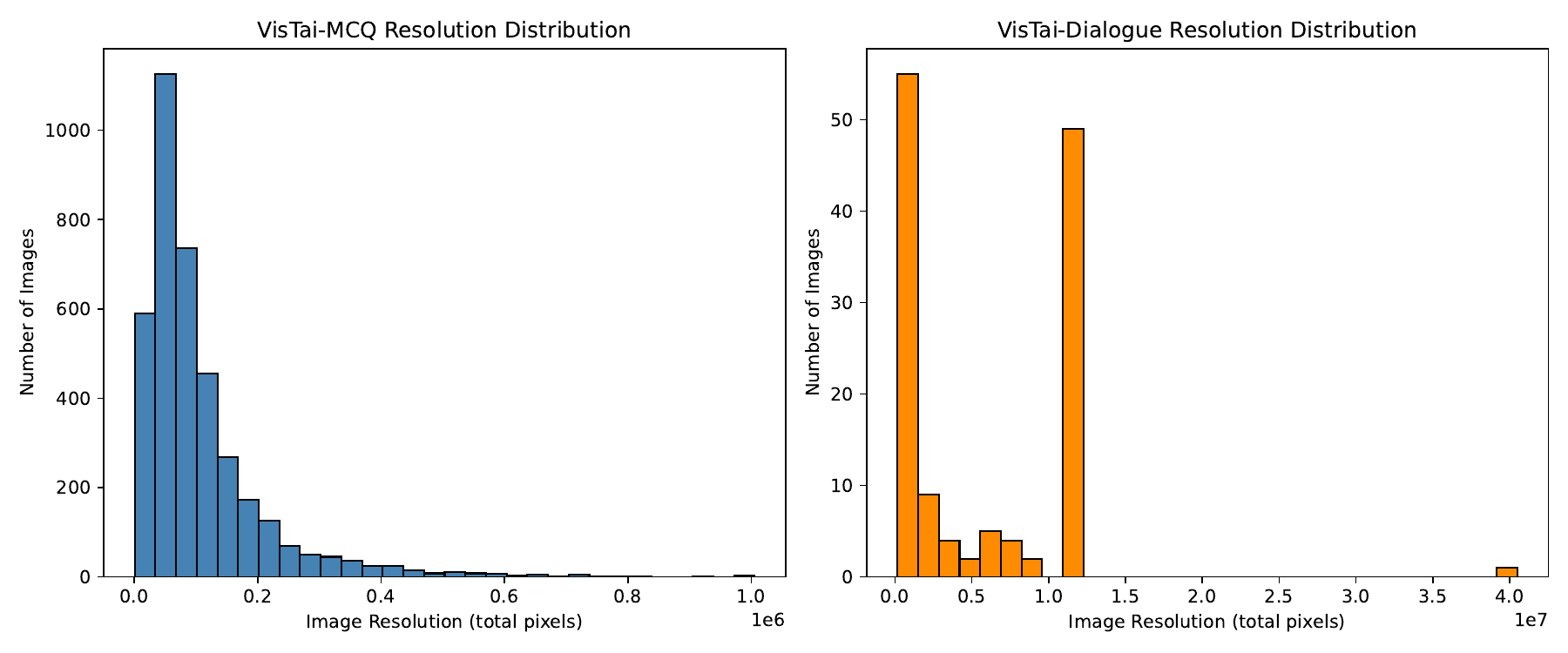}
    \caption{Comparison of image resolution distributions (in total pixels) for VisTW-MCQ (left) and VisTW-Dialogue (right). The x-axis shows the total number of pixels in each image, and the y-axis indicates how many images fall within each resolution range. The MCQ dataset tends to contain lower-resolution images, whereas the Dialogue dataset spans a broader range of resolutions.}
    \label{fig:resolution_distribution}
\end{figure}

Figure \ref{fig:aspect_ratio_distribution} (right) shows the aspect ratio of mostly long rectangle, most of the width is longer than the height. In the Figure \ref{fig:resolution_distribution} (right) for the MCQ dataset most of the image pixel is above 1M pixels.

\clearpage
\section{Benchmarked VLM Details}
\label{app:model_details}

For open weight models, we utilize local GPU resources (3090) for models under 11B parameters, while larger models are accessed through the Open Router API service. For closed weight models, we access them directly through each organization's official API.

\begin{table}[h]
\centering
\begin{tabular}{llll}
\toprule
\textbf{Model} & \textbf{Organization} & \textbf{Size} & \textbf{Arch} \\
\midrule
\multicolumn{4}{l}{\textit{Closed Source Models}} \\
\midrule
claude-3-haiku-20240307 \citep{claude3} & Anthropic & - & - \\
claude-3.5-sonnet-20241022 \citep{claude3.5} & Anthropic & - & - \\
gemini-1.5-flash \citep{geminiteam2024gemini15unlockingmultimodal} & Google & - & Dense \\
gemini-1.5-flash-8b \citep{geminiteam2024gemini15unlockingmultimodal} & Google & 8B & Dense \\
gemini-1.5-pro \citep{geminiteam2024gemini15unlockingmultimodal} & Google & - & MoE \\
gemini-2.0-pro-exp-02-05 \citep{geminiteam2024gemini15unlockingmultimodal} & Google & - & - \\
gemini-2.0-flash-thinking-exp-1219 \citep{geminiteam2024gemini15unlockingmultimodal} & Google & - & - \\
gemini-2.0-flash-001 \citep{geminiteam2024gemini15unlockingmultimodal} & Google & - & - \\
gemini-2.0-flash-lite-preview-02-05 \citep{geminiteam2024gemini15unlockingmultimodal} & Google & - & - \\
gpt-4o-2024-11-20 \citep{openai2024gpt4ocard} & OpenAI & - & - \\
gpt-4o-2024-08-06 \citep{openai2024gpt4ocard} & OpenAI & - & - \\
gpt-4o-mini-2024-07-18 \citep{openai2024gpt4ocard} & OpenAI & - & - \\
nova-lite-v1 & Amazon & - & - \\
\midrule
\multicolumn{4}{l}{\textit{Open Weights Models}} \\
\midrule
Gemma3 27B Instruct \citep{team2025gemma}& Google & 27.4B & Dense \\
Qwen2.5-VL-72b-instruct \citep{bai2025qwen25vltechnicalreport} & Alibaba & 72B & Dense \\
Qwen2.5-VL-7B-Instruct \citep{bai2025qwen25vltechnicalreport} & Alibaba & 7B & Dense \\
Qwen2-VL-72b-instruct \citep{bai2023qwenvlversatilevisionlanguagemodel} & Alibaba & 72B & Dense \\
Qwen2-VL-7B-Instruct \citep{bai2023qwenvlversatilevisionlanguagemodel} & Alibaba & 7B & Dense \\
InternVL2.5-8B \citep{chen2025expandingperformanceboundariesopensource} & OpenGVLab & 8B & Dense \\
InternVL2-8B-MPO \citep{chen2024fargpt4vclosinggap}  & OpenGVLab & 8B & Dense \\
InternVL2.5-4B \citep{chen2025expandingperformanceboundariesopensource} & OpenGVLab & 4B & Dense \\
InternVL2-8B 
 \citep{chen2024fargpt4vclosinggap} & OpenGVLab & 8B & Dense \\
InternVL2-4B 
 \citep{chen2024fargpt4vclosinggap} & OpenGVLab & 4B & Dense \\
InternVL2-2B \citep{chen2024fargpt4vclosinggap}  & OpenGVLab & 2B & Dense \\
nternVL2-1B \citep{chen2024fargpt4vclosinggap}  & OpenGVLab & 1B & Dense \\
Llama-3.2-90B-Vision-Instruct \citep{patterson2022carbonfootprintmachinelearning} 
 & Meta & 90B & Dense \\
Llama-3.2-11B-Vision-Instruct \citep{patterson2022carbonfootprintmachinelearning} & Meta & 11B & Dense \\
CogVLM2-llama3-chinese \citep{hong2024cogvlm2visuallanguagemodels} & THUDM & 19B & Dense \\
Breeze2-8B-Instruct \citep{research2025breeze2herdmodels} & MediaTek & 8B & Dense \\
Breeze2-3B-Instruct \citep{research2025breeze2herdmodels} & MediaTek & 3B & Dense \\
deepseek-vl2-tiny \citep{wu2024deepseekvl2mixtureofexpertsvisionlanguagemodels} & DeepSeek & - & MoE \\
deepseek-vl2-small \citep{wu2024deepseekvl2mixtureofexpertsvisionlanguagemodels} & DeepSeek & - & MoE \\
\bottomrule
\end{tabular}
\caption{Overview of evaluated models. For closed source models, sizes are marked with `-' where not publicly disclosed. Dense stands for Decoder only Dense Transformer architecture, MoE stands for Mixture of Experts architecture.}
\end{table}

\clearpage
\section{Prompts}
\label{app:vlm_prompt_section}

\subsection{VisTW-Dialogue Judgment Prompt}
\label{app:dialogue_judge_prompt}
\begin{figure}[h]
\begin{CJK}{UTF8}{bsmi}
\centering
\begin{AIboxC}{VisTW-Dialogue Example}
\includegraphics[width=0.4\textwidth, keepaspectratio]{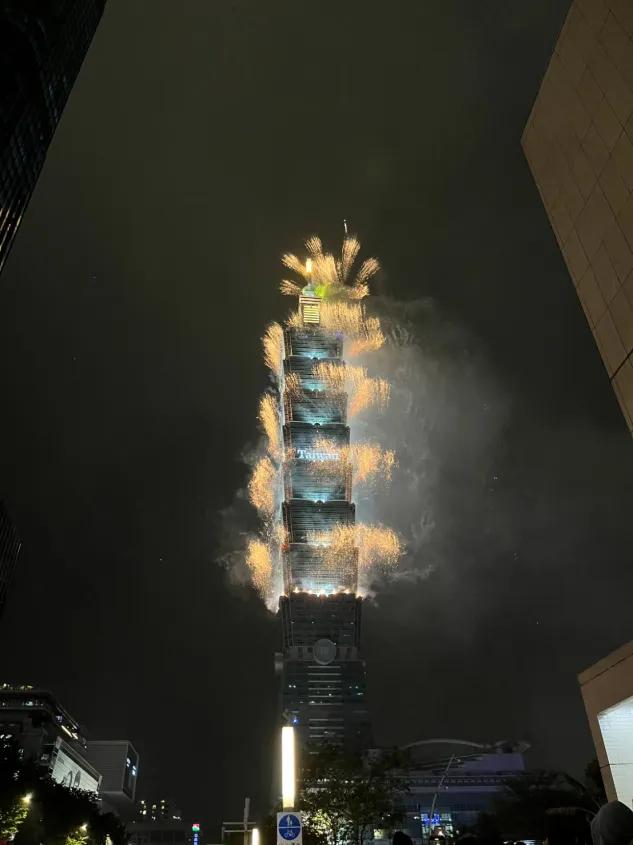} \\ \\
請根據使用者詢問的問題 [Question] 與正確答案 [Ground Truth], 去評價助手的回覆 [Assistant Response] 的評分，評分依照下方的評價指導手冊去評分。\\
\textcolor{blue}{Please evaluate the assistant's response based on the user's question [Question] and the correct answer [Ground Truth], and assess the score of the assistant's response [Assistant Response] according to the evaluation guidelines provided below.}\\ \\

\textnormal{[問題]} \textcolor{blue}{[Question]} \\
這在台灣哪的建築？能猜出這是什麼時間點拍的嗎？ \\
\textcolor{blue}{Where is this building located in Taiwan?\\Can the time period during which this photo was taken be determined?} \\

\textnormal{[評價助手的回覆]} \textcolor{blue}{[Assistant Response]} \\
這張照片拍攝的是台北101大樓，畫面中出現的煙火表演通常是在跨年時舉行的，因此可以推測這張照片大約是跨年夜所拍攝的。台北101在新年倒數時會舉行盛大的煙火秀，吸引了許多民眾前來觀賞。 \\
\textcolor{blue}{This photograph depicts Taipei 101, with fireworks display typically held during New Year's Eve celebrations. It is inferred that this photo was taken on New Year's Eve, as Taipei 101 annually hosts a grand fireworks show during the New Year countdown, drawing large crowds of spectators.} \\

\textnormal{[正確答案]} \textcolor{blue}{\textnormal{[Ground Truth]}} \\
這是台灣台北著名的「台北101」，此圖是元旦跨年時101的煙火秀 \\
\textcolor{blue}{This is the renowned Taipei 101 in Taipei, Taiwan. The image captures the New Year's Eve fireworks display at the building during the annual countdown celebration.} \\

你的回覆格式應該是如下: \textcolor{blue}{Your response should be in the format:} 
\begin{description}
    \item \textnormal{[解釋]:} (你的解釋) \textcolor{blue}{\textnormal{[Explanations]:} (Your explanations)}  \\
    \textcolor{red}{\textnormal{[Explanations]:} Completely correct}
    \item \textnormal{[評分]:} (int分數) \textcolor{blue}{\textnormal{[Scoring]:} (Int score)} \\
    \textcolor{red}{\textnormal{[Scoring]:} 10} 
\end{description}

\end{AIboxC}
\caption{VisTW-Dialogue for evaluating the response with prefect 10 points.}
\label{example: VisTW-Dialogue Example}
\end{CJK}
\end{figure}

Apart from providing judgement prompts Figure\ref{prompt:Judgement Prompt}, we utilize a 10 point scale for evaluation, with the score decreasing from 10. If there are any omissions or mistakes in the model explanations, appropriate deductions will be made. Besides, in Scoring Guidelines Figure\ref{prompt:Scoring Guidelines} we also explain the range definition represented by each score, so as to better evaluate the model's capabilities. In the Figure\ref{example: VisTW-Dialogue Example} example, according to the Scoring Method and Scoring Guidelines, the model explanations accurately aligned with the Ground Truth and addressed all answers with clear and structured language. This evaluation got a perfect score of 10.

\begin{figure}[t]
\begin{CJK}{UTF8}{bsmi}
\centering
\begin{AIboxC}{Judgement Prompt}
請根據使用者詢問的問題 [Question] 與正確答案 [Ground Truth], 去評價助手的回覆 [Assistant Response] 的評分，評分依照下方的評價指導手冊去評分。\\
\textcolor{blue}{Please evaluate the assistant's response based on the user's question [Question] and the correct answer [Ground Truth], and assess the score of the assistant's response [Assistant Response] according to the evaluation guidelines provided below.}\\
\textnormal{[問題]} \textcolor{blue}{[Question]}\\
\{question\} \\ 
\textnormal{[評價助手的回覆]} \textcolor{blue}{[Assistant Response]} \\
\{response\} \\ 
\textnormal{[正確答案]} \textcolor{blue}{\textnormal{[Ground Truth]}} \\
\{ground\_truth\} \\\\
\# 評分標註指南 \textcolor{blue}{\# Scoring Guidelines}\\
\{Scoring Guidelines\} \\
\begin{description}
    \item \#\# 評分方式 \textcolor{blue}{\textnormal{\#\# Scoring Method} }
    \begin{description}
        \item[1.] 首先將回答與標準答案比較 \textcolor{blue}{1. Responses were  compared with standard answers.}
        \item[2.] 評估以下幾點：\textcolor{blue}{2. Evaluate the following aspects:}
            \begin{description}
                \item[-] 是否準確？ \textcolor{blue}{- Whether the responses were accurate?}
                \item [-] 是否回答了問題的所有部分 \textcolor{blue}{- Whether all parts of the question were addressed?}
                \item [-] 是否清晰且有條理？ \textcolor{blue}{- Whether the responses were clear and well-structured?}
                \item [-] 是否提供有幫助的補充說明？ \textcolor{blue}{- Whether helpful explanations were provided?}
            \end{description}
        \item [3.] 從10分開始扣分：\textcolor{blue}{3. Points were deducted for errors from an initial 10-point score.}
            \begin{description}
              \item[-] 每個事實錯誤（-1至-2分）\textcolor{blue}{- Each factual error deducted 1–2 points.}
              \item[-] 遺漏資訊（-1至-2分）\textcolor{blue}{- Missing information deducted 1–2 points.}
              \item[-] 語言組織不佳（-1分）\textcolor{blue}{- Incoherent language structure deducted 1–2 points.}
              \item[-] 補充說明不當或錯誤（-1分）\textcolor{blue}{- Inappropriate or incorrect explanations deducted 1 point.}
            \end{description}
        \item [4.] 請簡短說明評分理由，包含：\textcolor{blue}{Scoring rationale includes concise explanations of:}
             \begin{description}
              \item[-] 做得好的地方\textcolor{blue}{- Well-addressed components or exceptional performance.}
              \item[-] 缺少或錯誤的部分\textcolor{blue}{- Missing or incorrect information}
              \item[-] 為何給予此分數\textcolor{blue}{- Why this score was awarded?}
            \end{description}
    \end{description}
\end{description} 

你的回覆格式應該是如下: \textcolor{blue}{Your response should be in the format:} 
\begin{description}
    \item \textnormal{[解釋]:} (你的解釋) \textcolor{blue}{\textnormal{[Explanations]:} (Your explanations)}  
    \item \textnormal{[評分]:} (int分數) \textcolor{blue}{\textnormal{[Scoring]:} (Int score)} 
\end{description}
\end{AIboxC}
\end{CJK}
\caption{Instruction prompt used in scoring the sample.}
\label{prompt:Judgement Prompt}
\end{figure}

\begin{figure}[h]
    \centering
    \begin{CJK}{UTF8}{bsmi}
    \begin{AIboxC}{Scoring Guidelines}
        \#\# 評分範圍（0-10分）\textcolor{blue}{Score Range (0 - 10 points)}
        \begin{description}
                 \item \#\#\# 10分：完美 \textcolor{blue}{10 points: Perfect}
                 \begin{description}
                     \item -  完全準確無誤 \textcolor{blue}{- Fully accurate and error-free}
                     \item -  回答問題的所有部分 \textcolor{blue}{- Address all parts of the questions}
                     \item -  清晰且條理分明 \textcolor{blue}{- Clear and well-structured}
                     \item -  提供有幫助的補充說明 \textcolor{blue}{- Helpful explanations}
                 \end{description}
                 \item \#\#\# 8 - 9分：非常好 \textcolor{blue}{8-9 points: Excellent}
                 \begin{description}
                     \item -  有些微錯誤或遺漏 \textcolor{blue}{- Minor errors or omissions}
                     \item -  主要重點都有涵蓋 \textcolor{blue}{- Key points are covered}
                     \item -  組織良好 \textcolor{blue}{- Clear and well-structured}
                     \item -  提供有用的細節說明 \textcolor{blue}{- Helpful explanations}
                 \end{description}
                 \item \#\#\# 6 - 7分：良好 \textcolor{blue}{6-7 points: Good}
                 \begin{description}
                     \item -  有一些小錯誤 \textcolor{blue}{- Some minor errors}
                     \item -  大部分重點都有提到 \textcolor{blue}{- Most key points are addressed}
                     \item -  組織尚可 \textcolor{blue}{- Adequate structure}
                     \item -  有一些相關說明 \textcolor{blue}{- Some relevant explanations}
                 \end{description}
                 \item \#\#\# 4 - 5分：普通 \textcolor{blue}{4-5 points: Average}
                 \begin{description}
                     \item -  有數個錯誤 \textcolor{blue}{- Several errors}
                     \item -  遺漏一些重點 \textcolor{blue}{- Some key points are omitted}
                     \item -  基本的組織結構 \textcolor{blue}{- Basic structure}
                     \item -  說明不夠完整 \textcolor{blue}{- Incomplete explanations}
                 \end{description}
                 \item \#\#\# 2 - 3分：不佳 \textcolor{blue}{2-3 points: Poor}
                 \begin{description}
                     \item -  有許多錯誤 \textcolor{blue}{- Numerous errors}
                     \item -  遺漏重要資訊 \textcolor{blue}{- Critical information is omitted}
                     \item -  組織不清楚 \textcolor{blue}{- Disorganized structure}
                     \item -  說明不正確或不足 \textcolor{blue}{- Incorrect or insufficient explanations}
                 \end{description}
                 \item \#\#\# 0 - 1分：不及格 \textcolor{blue}{0-1 points: Failing}
                 \begin{description}
                     \item -  大部分錯誤 \textcolor{blue}{- Mostly incorrect}
                     \item -  未回答問題重點 \textcolor{blue}{- The core question remains unanswered}
                    \item -  沒有明確的組織 \textcolor{blue}{- Absence of structure}
                     \item -  說明不相關或有誤導 \textcolor{blue}{- Irrelevant or misleading explanations}
                 \end{description}
             \end{description}
    \end{AIboxC}
    \end{CJK}
\caption{Scoring guidelines outline the criteria for each point}
\label{prompt:score_guideline}
\end{figure}















\label{prompt:no-risk-informed-moral-dispute}

\subsection{VisTW-MCQ Prompt}
\begin{figure}[h]
\begin{CJK}{UTF8}{bsmi}
\centering
\begin{AIboxC}{VisTW-MCQ - Zero-Shot COT Example}
\includegraphics[width=0.4\textwidth, keepaspectratio]{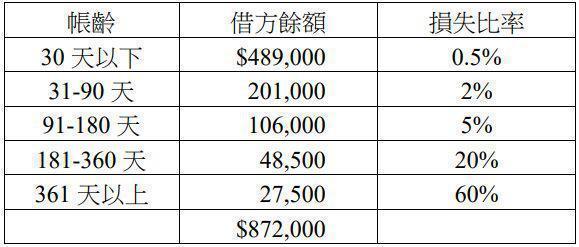} \\ \\
回答以下的多選題問題。並且在回覆的最後記得講格式： 答案: \$字母 而字母是 ABCDEFG 的其中一個。回答前請先一步一步(think step by step)想好答案。你必須使用中文回答。\\
    \textcolor{blue}{Answer the following multiple-choice questions. Additionally, remember to include the format at the end of your response: Answer: \$letter, where the letter is one of A, B, C, D, E, F, G. Before answering, carefully deliberate through the answer step by step. You must respond in Chinese.} \\ \\

[問題]: \\
甲公司係採用帳齡分析法估計預期信用減損損失，X8 年 12 月 31 日應收帳款帳齡分析資料如下：X8 年 5 月 12 日曾將確認無法收回之應收帳款\$35,680 沖銷；X8 年 12 月 31 日調整前備抵損失為借餘\$970。甲公司 X8 年度應認列之預期信用減損損失為何？ \\
\textcolor{blue}{
[Question]: \\
Company A uses the aging analysis method to estimate expected credit losses. The aging analysis data for accounts receivable as of December 31, Year X8 is as follows: On May 12, Year X8, the company wrote off accounts receivable of \$35,680 confirmed to be uncollectible; the allowance for credit losses had a debit balance of \$970 before adjustment on December 31, Year X8. What amount of expected credit loss should Company A recognize for the Year X8?
} \\
Ａ. \$37,965 \textcolor{blue}{Ａ. \$37,965}\\
Ｂ. \$36,935 \textcolor{blue}{Ｂ. \$36,935}\\
Ｃ. \$36,995 \textcolor{blue}{Ｃ. \$36,995}\\
Ｄ. \$38,935 \textcolor{blue}{Ｄ. \$38,935}\\

\textcolor{magenta}{
[COT]: \\
我將計算預期信用損失。我將每個帳齡類別的餘額乘以相應的損失率，然後將結果相加。最後，我會調整信用損失準備金中的借方餘額。
預期信用損失計算如下：
    1. 30天以下：\$489,000 * 0.5\% = \$2,445
    2. 31-90天：\$201,000 * 2\% = \$4,020
    3. 91-180天：\$106,000 * 5\% = \$5,300
    4. 181-360天：\$48,500 * 20\% = \$9,700
    5. 361天以上：\$27,500 * 60\% = \$16,500 \\
總預期信用損失：\$2,445 + \$4,020 + \$5,300 + \$9,700 + \$16,500 = \$37,965 \\
調整前信用損失準備的借方餘額：\$970
因此，應確認的預期信用損失金額為：\$37,965 + \$970 = \$38,935 \\
\textnormal{[答案]:} D}\\ \\
\textcolor{red}{
[COT]: \\
I will calculate the expected credit loss by multiplying the balance of each aging category by the corresponding loss rate and summing the results. Finally, adjustments are made for the debit balance in the credit loss reserve.
The expected credit loss calculation is as follows:
1. Under 30 days: \$489,000 × 0.5\% = \$2,445
2. 31-90 days: \$201,000 × 2\% = \$4,020
3. 91-180 days: \$106,000 × 5\% = \$5,300
4. 181-360 days: \$48,500 × 20\% = \$9,700
5. Over 361 days: \$27,500 × 60
Total expected credit loss: \$2,445 + \$4,020 + \$5,300 + \$9,700 + \$16,500 = \$37,965 \\
Debit balance of credit loss reserve before adjustment: \$970
Accordingly, the recognized expected credit loss amount is: \$37,965 + \$970 = \$38,935 \\
\textnormal{[Answer]:} D
}
\end{AIboxC}
\caption{VisTW-MCQ -  Zero-Shot COT with reasoning and correct response}
\label{example: VisTW-MCQ - Zero-Shot COT Example}
\end{CJK}
\end{figure}

\begin{figure}[h]
\begin{CJK}{UTF8}{bsmi}
\centering
\begin{AIboxC}{VisTW-MCQ - Direct Answer Example}
\includegraphics[width=0.6\textwidth, keepaspectratio]{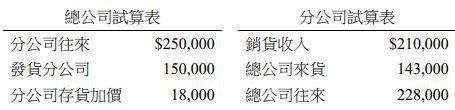} \\ \\
回答以下的多選題問題。並且在回覆的最後記得講格式： 答案: \$字母 而字母是 ABCD 的其中一個。直接給我答案回覆，Do not think step by step。你必須使用中文回答。\\
\textcolor{blue}{Answer the following multiple-choice question. Remember to include the format at the end of your response: Answer: \$letter, where the letter is one of A, B, C, or D. Provide only the answer in Chinese without any additional explanations. Do not think step by step. You must respond in Chinese.} \\ \\

[問題]: \\
X1 年間總公司總計將成本\$150,000 之商品運交分公司，轉撥價格均按成本加價 10\%，且分公司不自外界進貨。總公司運交分公司之一批商品，至期末分公司仍未收到，而導致「總公司往來」與「分公司往來」金額不一致。總公司與分公司於 X1 年底調整前試算表部分資料如下：試問分公司帳上期初存貨餘額為何？\\
\textcolor{blue}{
[Question]: \\
During the year X1, the Head Office transferred goods costing \$150,000 to the Branch at a transfer price of cost plus 10\%, with the Branch not sourcing inventory externally. A shipment of goods dispatched by the Head Office to the Branch remained undelivered at period-end, creating discrepancies between the balances of the Head Office Current Account and Branch Current Account. Selected pre-adjustment trial balance data from both entities as of year-end X1 are provided. \\
What was the Branch's beginning inventory balance?
} \\

Ａ. \$33,000 \textcolor{blue}{Ａ. \$33,000}\\
Ｂ. \$30,000 \textcolor{blue}{Ｂ. \$30,000}\\
Ｃ. \$0 \textcolor{blue}{Ｃ. \$0}\\
Ｄ. \$20,000 \textcolor{blue}{Ｄ. \$20,000}\\

\textcolor{red}{\textnormal{[答案]：}B}\\
\textcolor{red}{\textnormal{[Answer]:} B}
\end{AIboxC}
\caption{VisTW-MCQ - direct answer example with incorrect responses (the answer is A)}
\label{example:VisTW-MCQ - Direct Answer Exampl}
\end{CJK}
\end{figure}

In VisTW-MCQ, our prompts can be found in Figure\ref{prompt:VisTW-MCQ - Zero Shot COT Prompt} and Figure\ref{prompt: VisTW-MCQ - Direct Answer Prompt}
\label{app:VisTW-MCQ_prompt}. We divide the evaluation into two prompts: Zero-Shot COT and Direct-Answer to examine VLM model capabilities. In these examples of Figure\ref{example: VisTW-MCQ - Zero-Shot COT Example} and Figure\ref{example:VisTW-MCQ - Direct Answer Exampl}, we use red color to mark the model's responses and COT processes. For COT prompts, we specifically use "think step by step" to guide the model's reasoning further.

\begin{figure}
    \centering
    \begin{CJK}{UTF8}{bsmi}
    \begin{AIboxC}{VisTW-MCQ - Zero-Shot COT Prompt}
    回答以下的多選題問題。並且在回覆的最後記得講格式： 答案: \$字母 而字母是 ABCDEFG 的其中一個。回答前請先一步一步(think step by step)想好答案。你必須使用中文回答。\\
    \textcolor{blue}{Answer the following multiple-choice questions. Additionally, remember to include the format at the end of your response: Answer: \$letter, where the letter is one of A, B, C, D, E, F, G. Before answering, carefully deliberate through the answer step by step. You must respond in Chinese.}\\ \\ 
    \textnormal{[問題]:} \textcolor{blue}{[Question]}\\
    \{question\} \\ \\
    \textnormal{[COT]：}\textcolor{blue}{[COT:]}\\
    \{COT reasoning\} \\ \\
    \textnormal{[答案]：}\textcolor{blue}{[Answer:]}\\
    \{your answer\} \\ 
    \end{AIboxC}
    \caption{Instruction prompt used in VisTW-MCQ zero-shot COT}
    \label{prompt:VisTW-MCQ - Zero Shot COT Prompt}
   \end{CJK}
\end{figure}



\begin{figure}
    \centering
   \begin{CJK}{UTF8}{bsmi}
    \begin{AIboxC}{VisTW-MCQ - Direct-Answer Prompt}
    回答以下的多選題問題。並且在回覆的最後記得講格式： 答案: \$字母 而字母是 ABCD 的其中一個。直接給我答案回覆，Do not think step by step。你必須使用中文回答。\\
    \textcolor{blue}{Answer the following multiple-choice question. Remember to include the format at the end of your response: Answer: \$letter, where the letter is one of A, B, C, or D. Provide only the answer in Chinese without any additional explanations. Do not think step by step. You must respond in Chinese.}  \\ \\
     \textnormal{[問題]:} \textcolor{blue}{[Question]}\\
    \{question\} \\ \\
    \textnormal{[答案]：}\textcolor{blue}{[Answer:]}\\
    \{your answer\} \\ 
    \end{AIboxC}
    \caption{Instruction prompt used in VisTW-MCQ direct answer}
    \label{prompt: VisTW-MCQ - Direct Answer Prompt}
   \end{CJK}
\end{figure}


\section{Does exams with figure harder than normal exam questions?}
In this section we discuss whether exam questions grounded on visual context are harder than other questions from the same exam without context. Meaning does adding additional visual feature makes the benchmark harder?

\end{document}